\documentclass[runningheads]{llncs}

\usepackage{graphicx}
\usepackage{amsthm,amsmath,amssymb}

\begin{document}
\title{DAMO-NLP at NLPCC-2022 Task 2: Knowledge Enhanced Robust NER for Speech Entity Linking}
\titlerunning{DAMO-NLP at NLPCC-2022 Task 2}
\author{
Shen Huang\inst{1} \and
Yuchen Zhai\inst{2} \and
Xinwei Long\inst{3} \and
Yong Jiang\inst{1} \and 
Xiaobin Wang\inst{1} \and 
Yin Zhang\inst{2} \and 
Pengjun Xie\inst{1}\thanks{Corresponding author}
}
\authorrunning{S. Huang et al.}
\institute{
\inst{1}DAMO Academy, Alibaba Group \\
\email{\{pangda,yongjiang.jy,xuanjie.wxb,chengchen.xpj\}@alibaba-inc.com}\\
\inst{2}DMAC Group, DCD Lab, Zhejiang University\\
\email{\{zhaiyuchen,zhangyin98\}@zju.edu.cn}\\
\inst{3}University of Chinese Academy of Sciences\\
\email{longxinwei19@mails.ucas.ac.cn}
}
\maketitle              % typeset the header of the contribution
\begin{abstract}
Speech Entity Linking aims to recognize and disambiguate named entities in spoken languages. 
Conventional methods suffer gravely from the unfettered speech styles and the noisy transcripts generated by ASR systems.
In this paper, we propose a novel approach called Knowledge Enhanced Named Entity Recognition (KENER), which focuses on improving robustness through painlessly incorporating proper knowledge in the entity recognition stage and thus improving the overall performance of entity linking.
KENER first retrieves candidate entities for a sentence without mentions, and then utilizes the entity descriptions as extra information to help recognize mentions. The candidate entities retrieved by a dense retrieval module are especially useful when the input is short or noisy. 
Moreover, we investigate various data sampling strategies and design effective loss functions, in order to improve the quality of retrieved entities in both recognition and disambiguation stages.
Lastly, a linking with filtering module is applied as the final safeguard, making it possible to filter out wrongly-recognized mentions.
Our system achieves 1st place in Track 1 and 2nd place in Track 2 of NLPCC-2022 Shared Task 2.

\end{abstract}
\section{Introduction}

Speech Entity Linking (SEL), which aims to recognize and disambiguate named entities in spoken languages to a certain knowledge base (KB), is widely used in search engines, voice assistants and other speech analysis services. 
It's quite annoying but very common, for example, when you want news of Yao Ming who is the famous basketball player but the voice assistant returns you news of another Yao Ming who is a composer.
The spoken language, compared to the written language, is much more informal for its unfettered styles and high-frequency grammatical errors. All these bring great difficulties to SEL, not to mention the noisy transcripts generated by Automatic Speech Recognition (ASR) systems. 

The NLPCC 2022 Shared Task 2 provides researchers an opportunity to study Entity Linking (EL) problems with spoken languages and noisy texts. The task includes two tracks: (1) Track 1 - Entity Recognition and Disambiguation; (2) Track 2 - Entity Disambiguation-Only. The difference is that in Track 1 one needs to process the transcripts to extract named entities and disambiguate the extracted entities to the correct entry in a given knowledge base while in Track 2 gold standard named entities are given as input and all one needs to do is to disambiguate them. 

% Although many studies work on the joint framework

Previous work on end2end EL problem show that is extraction is harder than disambiguation. Following these lines of work, we perform some preliminary experiments and verify it in this competition. To tackle this problem, we propose a three-stage approach called Knowledge Enhanced Named Entity Recognition (KENER). Firstly it retrieves candidate entities from a KB for an input sentence without mentions via a dense retrieval module. Secondly the retrieved entity descriptions and the sentence contexts are utilized as extra information to help recognize mentions in the Named Entity Recognition (NER) module. Thirdly a linking with filtering module is applied to link the mentions to a KB while filtering out wrongly-recognized mentions. Finally, with smart sampling strategy and ensemble methods, our system achieves 1st place in track 1 and 2nd place in track 2 of NLPCC-2022 Shared Task 2.

Besides the system description, we make the following observations based on our experiments. \\
1. Our entity retrieval module can retrieve highly-related entities given a certain sentence without mentions. (Section \ref{CER}) \\
2. Our NER module can effectively utilize the candidate entities to improve the overall performance. (Section \ref{NER}) \\
3. Our linking module can tackle noisy mentions with the filtering mechanism . (Section \ref{Linking})

\section{Related Work} 
\noindent \textbf{Named Entity Recognition (NER)}: Most of the work takes NER as a sequence labeling problems and applies the linear-chain CRF~\cite{Lafferty2001}. Recently, the improvement of accuracy mainly benefits from stronger token representations such as pretrained contextual embeddings such as BERT~\cite{Devlin2019}, Flair~\cite{Akbik2019} and LUKE~\cite{Yamada2020}. Yu et al.~\cite{Yu2020} utilizes the strength of pretrained contextual embeddings over long-range dependency and encodes the document-level contexts for token representations. Wang et al.~\cite{Wang2021} proposes to mine external contexts of a sentence by retrieving to improve NER performance.

\noindent \textbf{Entity Disambiguation (ED)}: Most of the work focuses on contextual representation learning or features based on the entity mention~\cite{Rao2013}. DPR~\cite{Karpukhin2020} and BLINK~\cite{Wu2020} calculate the match between the mention and entity through a bi-encoder during retrieval and a cross-encoder is used for re-ranking~\cite{Wu2020}. Various context information is exploited, such as latent entity type information in the immediate context of the mention~\cite{Chen2020}.

\noindent \textbf{End to End EL}: There are some models that jointly processes NER and ED as well. Kolitsas et al.~\cite{Kolitsas2018} first proposes a neural end-to-end EL system that jointly discovers and links entities in a text document. Martins et al.~\cite{Martins2019} introduces a Stack-LSTM approach composed of SHIFT-REDUCE-OUT actions for multi-task learning. Cao et al.~\cite{Cao2021} creatively formulate entity linking as a sequence-to-sequence task, where an encoder-decoder architecture is used to autoregressively generate terms in a prefix tree of entity title vocabulary. Zhang et al.~\cite{Zhang2022} proposes the EntQA model that combines progress in entity linking with that in open-domain question answering and capitalizes on dense entity retrieval and reading comprehension.

\section{Methods}

\begin{figure}
\includegraphics[width=\textwidth]{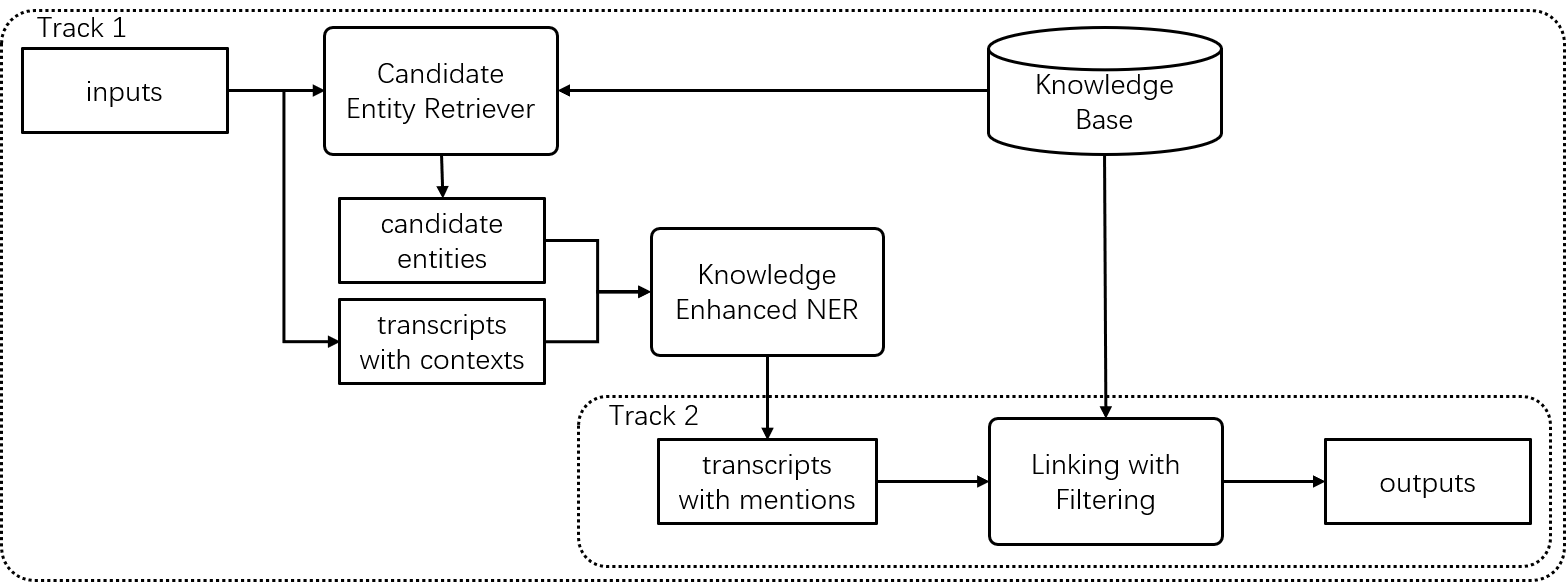}
\caption{Our proposed KENER system architecture for the shared task.} \label{fig1}
\end{figure}

Fig.~\ref{fig1} describes the architecture of our KENER system, which consists of three modules  sequentially: Candidate Entity Retriever, Knowledge Enhanced NER and Linking with Filtering. We utilize the same linking module for Track 1 and Track 2. 

\subsection{Candidate Entity Retriever} \label{CER}

The Candidate Entity Retriever module retrieves top-K candidate entities that might be mentioned in the input sentence. It adopts a variant of the bi-encoder structure, where the representation of an input sentence and a candidate entity from a KB are computed separately. The scores of candidate entities can then be calculated by the representations, so as to find the top-K candidate entities. We formulate the procedure as follows.

Given a sentence transcript $x \in \mathcal{V}^L$ and an entity $e \in \mathcal{E}$, where $\mathcal{V}$ denotes the vocabulary and $\mathcal{E}$ denotes the set of entities in a KB associated with text titles, aliases, and descriptions. $L$ is the length of a sentence.

The candidate entity retrieval score $\mathrm{score}_{r1}(x, e)$ can be computed by

\begin{equation*}
\begin{aligned}
S_1 &= \textbf{enc}_S([\mathrm{CLS}] ~x~ [\mathrm{SEP}] ~\mathrm{ctx(x)}~ [\mathrm{SEP}]) \\
E_1 &= \textbf{enc}_E([\mathrm{CLS}] ~\mathrm{title(e)}~ [\mathrm{SEP}] ~\mathrm{alias(e)}~\mathrm{desc(e)}~ [\mathrm{SEP}]) \\
\mathrm{score}_{r1}(x, e) &= S_1^T E_1
\end{aligned}
\end{equation*}

\noindent where the $\mathrm{ctx}(x)$ denotes the context sentences of $x$ in a speech and $\mathrm{title(e)}$, $\mathrm{alias(e)}$, $\mathrm{desc(e)}$ denote the text title, alias and description of the entity $e$.

During inference time, we use Faiss~\cite{Faiss} with precomputed $E$ for each $e \in \mathcal{E}$ for fast retrieval. 

\noindent \textbf{Multi-label NCE Loss.} 
For the reason that one sentence may contain multiple entities, we train the retriever using a multi-label variant of noise contrastive estimation (NCE) following~\cite{Zhang2022}. The optimization objective is
$$
\mathrm{max} \sum_{e \in \mathcal{E}(x)} \mathrm{log} (\frac{\mathrm{exp}(\mathrm{score}_{r1}(x, e))}{\mathrm{exp}(\mathrm{score}_{r1}(x, e)) + \sum_{e' \in \mathrm{N}(\mathcal{E}, x)}\mathrm{exp}(\mathrm{score}_{r1}(x, e'))})
$$
\noindent where $\mathrm{N}(\mathcal{E}, x) \subset \mathcal{E} \setminus \mathcal{E}(x) $ is a set of negative examples that excludes all gold entities $\mathcal{E}(x)$.

\noindent \textbf{Hard Negative Sampling.} To retrieve entities of high quality, the hard negative sampling strategy is adopted. Instead of randomly sampling negative examples, we use the model from the last iteration to generate negative examples. It has been proved that the high-score negative examples can significantly help polish up the model. For this shared task, the negative examples are sampled randomly for the first iteration, while hard negative examples are mined for the next two iterations.

\subsection{Knowledge Enhanced NER} \label{NER}

The Knowledge Enhanced NER module extracts entity mentions for a sentence. In addition to the sentence itself, the contexts of the sentence in a speech and the retrieved candidate entities are also used as input. We also tried to use the audio but no significant improvement was revealed. We formulate NER as a sequence labeling task and the classical model BERT-CRF is adopted in our method. 

\begin{figure}
\includegraphics[width=\textwidth]{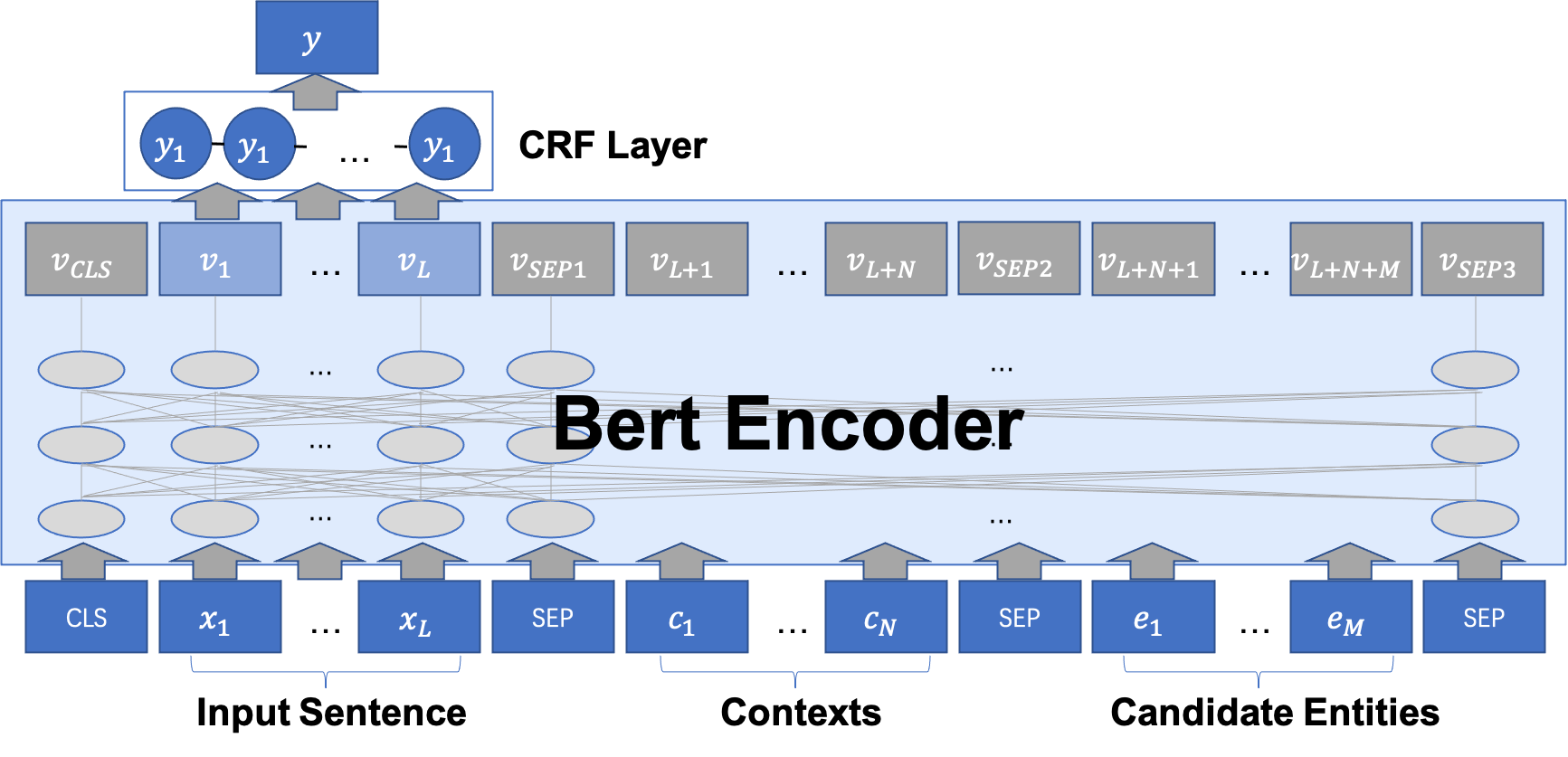}
\caption{BERT-CRF model with extra context and candidate entities as input.} \label{fig2}
\end{figure}

As depicted in Fig.~\ref{fig2}, we simply concatenate a sentence, its context and retrieved candidate entities (using its title, aliases and descriptions) as input to the BERT encoder:
$$
\hat{x} = [\mathrm{CLS}] ~x~ [\mathrm{SEP}] ~\mathrm{ctx}(x)~ [\mathrm{SEP}] ~\mathcal{\hat{E}}_1(x)~\mathcal{\hat{E}}_2(x)~...~\mathcal{\hat{E}}_M(x)~ [\mathrm{SEP}]
$$
where $\mathcal{\hat{E}}_i(x) = \mathrm{title}(\mathcal{E}_i(x))~\mathrm{alias}(\mathcal{E}_i(x))~\mathrm{desc}(\mathcal{E}_i(x))$ denotes the text representation of the $i$th retrieved candidate entity of $x$.

Experimental results prove that simple method works as well. The extra information brought by contexts and candidate entities can help generate better sentence representation. Then we drop other embeddings from contexts or candidate entities, and only keep the raw sentence embeddings for decoding. A CRF layer is used as the decoder to model the transition of tags.

\subsection{Linking with Filtering} \label{Linking}

A common linking module focuses on linking and disambiguating the given mentions to entities from a KB. In addition to that, our proposed Linking with Filtering module also tries to solve the problem when the given mentions are noisy and might contain errors. 

Following the mainstream framework, a bi-encoder is used for retrieval and a cross-encoder is used for disambiguation or ranking. The bi-encoder independently embeds the mention context and the entity descriptions so candidate entities can be retrieved in a dense space.
Each candidate is then examined more carefully with a cross-encoder, that concatenates the mention and entity text for encoding their correlation.

Two additional entities are added to the provided KB: $\mathrm{NIL}$ and $\mathrm{ERROR}$. $\mathrm{NIL}$ is used for those mentions that are actually named entities but cannot be found in the given KB. $\mathrm{ERROR}$ is used for some other mentions that are not named entities. This is designed to filter out the possible prediction errors generated by NER modules.
As gold mentions are given, $\mathrm{ERROR}$ is not applied in Track 2.

\noindent \textbf{Hard Negative Sampling.} We also used the same hard negative sampling strategy as we do in Section~\ref{CER} for training the bi-encoder, despite of their inputs for the encoder differ.

\noindent \textbf{List-wise Loss.} Instead of point-wise or pair-wise losses, we adopt the list-wise loss for training the cross-encoder, which computes the Kullback-Leibler divergence (KL divergence) to explicitly measure the difference between the predicted and target distributions.

\noindent \textbf{Dynamic Sampling.} Aiming to reduce the size of candidates for disambiguation while keeping the candidate entities hard to disambiguate, we propose a dynamic sampling strategy that samples candidates with different probabilities. The higher retrieval score one candidate gets, the higher probability it is sampled.

\subsection{Ensemble Methods}

Ensemble methods are used twice in the end2end entity linking procedure. The first time it is used to gather NER models to generate better results and the second time it works for the disambiguation models to re-rank the candidate entities.

We ensemble the NER models in two strategies using voting. One is for better F1 score and the other is for better recall. We count the predicted tags for each token from all models and determine the final tag by the distribution. The F1 strategy takes ``O'' as a normal tag when voting wile the recall strategy sets a threshold for ``O'' so that it's less probable to predict a ``O'' tag. With our proposed Linking with Filtering module, there's no need to worry about the noise.

The linking models are ensembled in a hybrid way. Not only the last ranking scores are used, but also the retrieval scores are taken into consideration. We re-rank the candidates with the weighted sum of retrieval and ranking scores and obtain the final linking results.

\section{Experiments}

\subsection{Data Analysis and Experimental Settings}
The task provides annotated training data, unlabeled test data and the corresponding audios. All corpus provided are obtained from TED talks\footnote{https://www.ted.com/talks}. It also provides a knowledge base from Wikidata\footnote{http://www.wikidata.org/}.

Track 1 and Track 2 share a training set consisting of 1936 talks. The statistics of the training and test data is shown in Table~\ref{tab1}.

\begin{table}[!htbp]
\caption{Dataset statistics.}\label{tab1}
\centering
\setlength\tabcolsep{5pt}
\begin{tabular}{c|c|c|c}
\hline
Data & sentence number & avg. sentence length & avg. mention number \\
\hline
train & 223348 & 32.16 & 1.40 \\
track1 test & 2905 & 37.99 & - \\
track2 test & 2923 & 38.64 & 1.49 \\
\hline
\end{tabular}
\end{table}

No validation set is provided. So we manually split the full training set into 4:1 for training and validation. The tuning and ablation studies are all conducted on our validation set.

The provided knowledge base has a high recall of 95\%. We check the rest 5\% entities which are linked to NIL and find their corresponding wiki pages don't exists veritably. So the official knowledge base is the only KB used in our system.

The main pretrained models we use for NER and ED are RoBERTa-large~\footnote{https://huggingface.co/roberta-large} and co-condenser-wiki~\footnote{https://huggingface.co/Luyu/co-condenser-wiki} respectively.

In our NER model, the top-16 candidate entities are utilized as extra context for it achieves a relatively high recall of 93\% and the transformer structure cannot handle too long inputs.

\subsection{Main Results}

The evaluation results on test sets of Track 1 and Track 2 are released by the organizer. As shown as Table~\ref{tab2} and Table~\ref{tab3}, our system achieves 1st place in Track 1 and 2nd place in Track 2 of NLPCC-2022 Shared Task 2. 

As we can see that, the performance of our system in Track 2 is comparable (-0.16pt) with the 1st place system. However in Track 1 of Entity Recognition and Disambiguation, we beat the 2nd and 3rd place system by 2.9pt and 10.7pt, which proves the advantage of our proposed KENER method. Due to the task difference of Track 1 and Track 2, we can conclude that this advantage stems from the knowledge introduced in the NER stage.

\begin{table}
\caption{Track1 submission results.}
\label{tab2}
\centering
\setlength\tabcolsep{10pt}
\begin{tabular}{l|c}
\hline
System Name & F1 \\
\hline
Ours & 0.7460 \\
HITsz\_TMG & 0.7172 \\
xmu-nlp-team & 0.6390 \\
\hline
\end{tabular}
\end{table}

\begin{table}
\caption{Track2 submission results.}
\label{tab3}
\centering
\setlength\tabcolsep{10pt}
\begin{tabular}{l|c}
\hline
System Name & F1 \\
\hline
HITsz\_TMG & 0.8884 \\
Ours & 0.8868 \\
\hline
\end{tabular}
\end{table}

\subsection{Ablation Study}
We conduct three ablation studies to validate the effectiveness of individual components of our framework. Concretely, we first investigate the influence of retrieval methods. The results are shown in Table~\ref{tab4}, where stage 1 and stage 3 denote retrieval without mentions and with mentions respectively. We can see that: (1) The hard negative mining is critical for both the candidate entity retriever and the entity linking. The iterative hard negatives can lead to further improvements. (2) It is important to excluding gold entities in the normalization term of NCE, which leads to a massive increase in Recall@1 (38.44).

To further understand how the retrieval module affects our framework, we investigate the NER performances for different retrieval-augmented strategies. We train and evaluate the model with various contexts. The results are shown in Table~\ref{tab5}. We can observe that all these contexts leads to significantly improvements, especially for the external TED context. This indicates that the document-level information is more important for NER.

Table~\ref{tab6} shows an ablation study for the entity linking module. We report the Accuracy on the validation set of the Track 2 dataset. We find that the Hybrid Ensemble method helps boost the results by a large margin, and the List-wise Loss and the Dynamic Sampling Strategy brings additional gains. This experiment result shows that the retriever is critical for the final re-ranking performance.

\begin{table}[!htbp]
\caption{Retrieval experiments.}\label{tab4}
\centering
\setlength\tabcolsep{5pt}
\begin{tabular}{c|l|c|c|c|c|c}
\hline
Stage & Methods & Rec@1 & Rec@16 & Rec@32 & Rec@64 & Rec@128 \\
\hline
1 & Baseline & 25.34 & 53.91 & 90.16 & 95.64 & 96.82 \\
1 & w/ Multi-label NCE Loss & 38.44 & 62.19 & 92.22 & 95.89 & 97.04 \\
1 & w/ Hard Negative (HN) & 59.60 & 93.24 & 95.11 & 96.32 & 97.66 \\
\hline
3 & Baseline & 56.62 & 91.73 & 97.42 & 98.92 & 99.12 \\
3 & w/ First Stage HN & 82.19 & 95.92 & 98.14 & 99.16 & 99.35 \\
3 & w/ Second Stage HN & 83.23 & 97.65 & 98.54 & 99.31 & 99.47 \\
\hline
\end{tabular}
\end{table}

\begin{table}[!htbp]
\caption{NER experiments.}\label{tab5}
\centering
\setlength\tabcolsep{10pt}
\begin{tabular}{l|c}
\hline
Methods & F1 \\
\hline
Ours & 70.31 \\
w/o Context & 69.63 \\
w/o Hard Negative & 68.19 \\
w/o Candidate Entities & 68.32 \\
w/o Context \& Candidate Entities & 67.21 \\
\hline
w/ Ensemble & 71.09 \\
\hline
\end{tabular}
\end{table}

\begin{table}[!htbp]
\caption{Disambiguation experiments.}\label{tab6}
\centering
\setlength\tabcolsep{10pt}
\begin{tabular}{l|c}
\hline
Methods & F1 \\
\hline
Ours & 85.71 \\
w/o List-wise Loss & 85.33 \\
w/o Dynamic Sampling & 85.15 \\
\hline
w/ Hybrid Ensemble & 86.98 \\
\hline
\end{tabular}
\end{table}
% w/o Filtering & \#TODO \\

\section{Conclusions}
In this paper, we propose a novel knowledge enhanced method called KENER, which aims to leverage knowledge in the NER phase to improve the overall performance of entity linking. It consists of three-stage modules, each focusing on one question we raise in Section 1. We attribute KENER's outperformance to the following reasons: (1) The hard-negative sampling strategy and multi-label NCE loss can contribute to better retrieval results. (2) The NER model benefits greatly from the retrieved candidate entities. (3) The filtering mechanism is good when the input contains noisy mentions. There are also scopes for improvement in our approach. Currently the three stages are trained separately, which ignores the internal correlation among them. We are looking forward to a unified solution that formulates the NER and ED in one framework.


\end{document}